\def\year{2022}\relax

\documentclass[letterpaper]{article}
\usepackage{aaai22}
\usepackage{times}
\usepackage{helvet}
\usepackage{courier}
\usepackage[hyphens]{url}
\usepackage{graphicx} 
\usepackage{natbib}
\usepackage{caption}
\DeclareCaptionStyle{ruled}{labelfont=normalfont,labelsep=colon,strut=off}
\usepackage{amsmath,amssymb,amsthm}
\usepackage{enumerate}

\newtheorem{definition}{Definition}
\newtheorem{theorem}[definition]{Theorem}

\usepackage[noend,linesnumbered,ruled]{algorithm2e}
\usepackage{booktabs}
\usepackage{amsmath}
\usepackage{amsfonts}
\usepackage{dsfont}
\usepackage{bbold}

\newcommand{\inlinecite}[1]{\citeauthor{#1}~(\citeyear{#1})}

\newcommand{\np}{\ensuremath{\mathbf{NP}}}

\newcommand{\reals}{\mathbb{R}}
\newcommand{\vstar}{V^{*}}

\newcommand{\bracket}[1]{[\![#1]\!]}

\newcommand{\modelsum}{\textsc{\small GNN-Sum}\xspace}
\newcommand{\modelmax}{\textsc{\small GNN-Max}\xspace}
\newcommand{\pgoal}[1]{#1_{G}}
\newcommand{\ppos}[2]{#1^{\mathbb{#2}}}
\newcommand{\ptarget}{\textsc{\small T}\xspace}
\newcommand{\pedge}{\textsc{\small E}\xspace}
\newcommand{\ppath}{\textsc{\small P}\xspace}
\newcommand{\pspath}{\textsc{\small SP}\xspace}
\newcommand{\conn}{\textsc{\small CONN}\xspace}

\newcommand{\multiset}[1]{\{\!\!\{ #1 \}\!\!\}}
\newcommand{\A}{\mathfrak{D}}
\newcommand{\R}{\mathcal{\cal R}}

\newcommand{\agg}{\mathbf{agg}\xspace}
\newcommand{\comb}{\mathbf{comb}\xspace}

\newcommand{\CLS}{\text{CLS}}
\newcommand{\C}{\mathsf{C}}
\newcommand{\GC}{\mathsf{GC}}
\newcommand{\FO}{\mathsf{FO}}
\renewcommand{\L}{\mathcal{L}}

\newcommand{\mlp}{\mathbf{MLP}\xspace}
\newcommand{\pclear}{\textsc{\small Clear}\xspace}
\newcommand{\pon}{\textsc{\small On}\xspace}
\newcommand{\pholding}{\textsc{\small Holding}\xspace}

\newcommand{\pvisited}{\textsc{\small Visited}\xspace}
\newcommand{\patrobot}{\textsc{\small At-robot}\xspace}
\newcommand{\pconnected}{\textsc{\small Connected}\xspace}
\newcommand{\pat}{\textsc{\small At}\xspace}
\newcommand{\patrobby}{\textsc{\small At-robby}\xspace}

\newcommand{\pgripper}{\textsc{\small Gripper}\xspace}
\newcommand{\pfree}{\textsc{\small Free}\xspace}
\newcommand{\patsoilsample}{\textsc{\small At-soil-sample}\xspace}
\newcommand{\pcantraverse}{\textsc{\small Can-traverse}\xspace}
\newcommand{\proad}{\textsc{\small Road}\xspace}
\newcommand{\pvehicle}{\textsc{\small Vehicle}\xspace}
\newcommand{\pin}{\textsc{\small In}\xspace}
\newcommand{\pcapacity}{\textsc{\small Capacity}\xspace}
\newcommand{\pcapacitypredecessor}{\textsc{\small Capacity-Predecessor}\xspace}
\newcommand{\pdirty}{\textsc{\small Dirty}\xspace}
\newcommand{\padjacent}{\textsc{\small Adjacent}\xspace}

\newcommand{\tup}[1]{\langle #1 \rangle}

\newcommand{\citeay}[1]{\citeauthor{#1} (\citeyear{#1})}

\newcommand{\mminus}{\hspace{-.05em}\raisebox{.15ex}{\footnotesize$\downarrow$}}

\newcommand{\GT}[1]{#1{\,>\,}0}

\newcommand{\DEC}[1]{#1\mminus}

\newcommand{\prule}[2]{\{ #1 \} \mapsto \{ #2 \}}

\newcommand{\Q}{\mathcal{Q}}

\title{Learning General Optimal Policies with Graph Neural Networks: \\Expressive Power, Transparency, and Limits}
\author{
Simon St\r{a}hlberg,\textsuperscript{\rm 1}
    Blai Bonet,\textsuperscript{\rm 2}
    Hector Geffner\textsuperscript{\rm 2,3,1} \\
}
\affiliations{
\textsuperscript{\rm 1}Link\"{o}ping University, Sweden \\
    \textsuperscript{\rm 2}Universitat Pompeu Fabra, Spain \\
    \textsuperscript{\rm 3}Instituci\'o Catalana de Recerca i Estudis Avan\c{c}ats (ICREA), Barcelona, Spain \\
    simon.stahlberg@liu.se, bonetblai@gmail.com, hector.geffner@upf.edu
}

\nocopyright
\hypersetup{nolinks=true}

\begin{document}
\allowdisplaybreaks

\maketitle

\begin{abstract}
  It has been recently shown that general policies for many classical planning
  domains can be expressed and learned in terms of a pool of features defined
  from the domain predicates using a description logic grammar.
  At the same time, most description logics correspond
  to a fragment of $k$-variable counting logic ($\C_k$) for $k=2$, that has
  been shown to provide a tight characterization of the expressive power of
  graph neural networks.
  In this work, we make use of these results to understand the power and limits
  of using graph neural networks (GNNs) for learning optimal general policies
  over a number of tractable planning domains where such policies are known
  to exist.
  For this, we train a simple GNN in a supervised manner to approximate the
  optimal value function $\vstar{}(s)$ of a number of sample states $s$.
  As predicted by the theory, it is observed that general optimal policies
  are obtained in domains where general optimal value functions can be defined
  with $\C_2$ features but not in those requiring more expressive $\C_3$ features.
  In addition, it is observed that the features learned are in close correspondence
  with the features needed to express $\vstar{}$ in closed form.
  The theory and the analysis of the domains let us understand the features
  that are actually learned as well as those that cannot be learned in this way,
  and let us move in a principled manner from a combinatorial
  optimization approach to learning general policies to a potentially, more
  robust and scalable approach based on deep learning.
\end{abstract}

\section{Introduction}

Deep learning (DL) and deep reinforcement learning (DRL) are behind most of  key milestones in AI of recent years
 \cite{dqn,lecun2015deep,silver2,silver2017mastering}.
Yet, these methods struggle to produce solutions that are structurally general   \cite{bengio:high-level}. Even in simple tasks,
such as retrieving a key to open a door in a simple  environment, they may require a large number of simulations, and even
then, they may fail to generalize to all possible situations \cite{babyAI}. Interestingly, the computation of general policies
has  been  addressed recently in a model-based setting that assumes that a general model of the actions is known
in terms of action schemas and predicates  \cite{bonet:ijcai2018,frances:aaai2021}. This paper  is a  step aimed at bringing  these threads together  with two motivations:  to replace the combinatorial methods that
have been proposed to learn general policies by more robust and scalable deep learning methods, and to
do so in a principled manner where the intermediate representations and  experimental results,  both positive and negative,
can be understood.

For this, we exploit two  existing results. On the one hand, the realization that  general policies and value functions
for many classical benchmark  domains can be expressed  in terms  of features defined from the domain predicates
using a description logic (DL) grammar \cite{martin:kr2000,fern:general,bonet:aaai2019,frances:ijcai2019}. On the other, the correspondence established
between the   expressive power of a decidable  fragment of first-order logic, called $\C_2$,  which includes
most common  DLs \cite{description-logics}, and the expressive power of graph neural networks (GNNs)
\cite{barcelo:gnn,grohe:gnn}. The two results together suggest that   general policies
could  be learned  from the domain predicates  directly by  means of GNNs, except for those  which are not expressible in terms of $\C_2$ features at all.

In this paper, we carry out this exploration in a  context where the learned general policies are expected to  be \emph{optimal},
leading to   optimal (shortest)  plans in  any  instance of the target class of problems $\Q$.
In addition, instead of seeking for representations  of an  optimal policy, we seek  representations
of the  optimal value function. If this function $V$ is optimal, the  policy $\pi_V$ greedy in $V$ is
optimal as well. The focus on optimal values allows us to learn the general  function $V$ using labeled data
in  the form of  pairs $\tup{s,V^*(s)}$, and to evaluate the learned function $V$ in a crisp  manner
where  \emph{the execution of a non-optimal transition at any state is an error.} The optimality requirement is thus a
methodological choice which simplifies the training and evaluation procedures  in order to determine whether the
graph neural networks manage to  learn the value functions that can be expressed in terms of  $\C_2$ features
without having to make explicit the feature pool or the  underlying  grammar.
Recent works have used deep learning methods for  addressing  similar problems in the broader setting of stochastic MDPs \cite{sylvie:asnet,mausam:dl}.
Our approach is inspired by them and follows on their footsteps, but
it is not so much focused on performance but on understanding the scope of the methods
and the features learned.

The rest of the paper is organized as follows. We review classical planning,
general policies and value functions, and present value functions
for a number of tasks in terms of logical features, most in $\C_2$.
We review GNNs, their relation to finite-variable logics, and
define the architecture used for learning value functions.
We report the experiments and analyze results, discuss related work, and conclude.

\section{Classical Planning}

A classical planning instance is a pair  $P\,{=}\,\tup{D,I}$ where
$D$ is a  first-order planning \emph{domain} and $I$ is \emph{instance}  information \cite{geffner:book,ghallab:book}.
The  planning domain $D$ contains a set of predicate symbols $p$ and a set of  action schemas with preconditions
and effects given by atoms $p(x_1, \ldots, x_k)$ where
each $x_i$ is an argument of the schema.
The instance information is a  tuple  $I\,{=}\,\tup{O,Init,Goal}$   where $O$ is a (finite) set of object
names $c_i$, and $Init$ and $Goal$ are  sets  of \emph{ground atoms} $p(c_1, \ldots, c_k)$.
This is   the structure of planning problems  expressed in  PDDL \cite{pddl:book}
where  the domain and instance  information are  provided in separate files.

A classical problem $P\,{=}\,\tup{D,I}$ encodes  a  state model $S(P)=\tup{S,s_0,S_G,Act,A,f}$
in compact form where the states $s \in S$ are sets of ground atoms from $P$,
$s_0$ is the initial state $I$, $S_G$ is the set of goal  states $s$ such that $S_G \subseteq s$,
$Act$ is the set of ground actions in $P$, $A(s)$ is the set of ground actions
whose preconditions are (true) in $s$, and $f$ is the transition function
so that $f(a,s)$ for $a \in A(s)$ represents the state $s'$ that follows
action $a$ in the state $s$.   An action sequence $a_0, \ldots, a_{n}$
is applicable in $P$ if $a_i \in A(s_i)$ and  $s_{i+1}=f(a_i,s_i)$, for
$i=0, \ldots, n$, and it is a plan if $s_{n+1} \in S_G$.
The \emph{cost}  of  a plan is   assumed  to be given by its length
and  a plan is \emph{optimal} if there is no shorter plan.

The  representation of  planning problems $P$ in two parts $D$ and $I$, one that is general,
and the other that is specific, is essential for defining and computing general
policies, as  the instances are assumed to come all from the same domain.  Recent work has addressed the problem of learning such first-order representations
from unstructured data   \cite{asai:fol,bonet:ecai2020,ivan:kr2021}.

\section{General Policies and Value Functions}

Generalized planning  studies the representation
and computation of  policies that solve many classical
planning instances from the same domain at once \cite{srivastava08learning,bonet:icaps2009,hu:generalized,BelleL16}.

For example,  $\Q_{clear}$  consists of all classical problems in Blocksworld where a block $x$ must
be cleared, regardless of the number or initial configuration of blocks,
and a  general policy for  $\Q_{clear}$ can be expressed in terms of  the two  features $\Phi=\{H,n\}$,
where $H$ is a true in a state if a block is being held, and $n$ represents the number of blocks above
the target block $x$, by means of  the rules  $\prule{\neg H, \GT{n}}{H, \DEC{n}}$
and $\prule{H}{\neg H}$ \cite{bonet:ijcai2018}.
The first rule  says that  when the gripper is empty and there are blocks above $x$,
any action that decreases $n$ and makes $H$ true should be selected.
The second that when the gripper is not empty, any action that  makes $H$ false and does not affect $n$ should be selected.
It has been shown that general policies of this  form can  be learned without supervision
by solving a Max-Weighted SAT theory $T({\cal S},{\cal F}$)
where ${\cal S}$ is a set of sampled state transitions, and $\cal F$ is a large but finite pool of
Boolean and numerical features obtained from the domain predicates \cite{frances:aaai2021}.

In this work, it  is convenient to  represent policies in terms of  \emph{value functions}.
As it is usual in dynamic programming and RL  \cite{sutton:book,bertsekas:dp}, a value function $V$ defines a (non-deterministic) \emph{greedy policy} $\pi_V$ that selects in a state $s$ \emph{any}
possible successor state $s'$ with minimum $V(s')$ value, under the assumption that actions are deterministic and of the same
cost.  A  policy $\pi$ solves an instance $P$ if the state transitions compatible with $\pi$,  starting with the initial state,
always end up in a  goal state, and   $\pi$ solves a class of problems $\Q$ if it solves each problem in the class.
Similarly, $\pi$ solves $P$ and $\Q$ \emph{optimally} when  goals are always reached optimally. Clearly, if $V$ is optimal, i.e.,\ $V$ is the optimal cost function $V^*$, the greedy policy $\pi_V$ is optimal too.
The general value functions are defined over general features $\phi_i$,
which are well-defined state functions over the states arising in instances of $\Q$ as:
\begin{alignat*}{1}
  V(s) = F(\phi_1(s), \ldots, \phi_k(s))
\end{alignat*}
where  $\phi_i(s)$ is the value of the feature $\phi_i$  in state $s$.
Value functions that are \emph{linear} have the form:
\begin{alignat*}{1}
  V(s)  = \textstyle \sum_{1\leq i\leq k}  w_i \phi_i(s)
\end{alignat*}
\noindent where the coefficients $w_i$ are constants that do not depend on the states.
For example, the general value function for the collection of problems $\Q_{clear}$, assuming different actions for picking and placing objects, is:
\begin{alignat*}{1}
  V = 2 n + H
\end{alignat*}
where the states  are left implicit, and the Boolean feature $H$ is assumed to have value $1$ when true, and
$0$ otherwise (the opposite for $\neg H$). This value function is optimal for $\Q_{clear}$. In planning, two types of
linear value functions that have been used are ``potential heuristics'' \cite{potential-heuristics},
that are instance-dependent and use   features that  stand for conjunction of atoms, and
``generalized potential heuristics'' \cite{frances:ijcai2019}  that  use the Boolean and the numerical  description logic  features introduced by
\citeay{bonet:aaai2019}.

\subsection{Domain Predicates, Features, and Logics}

A key problem in reinforcement learning (RL), and in particular in RL with  linear function approximation,
is the choice of the features \cite{givan:features,rl:lfa,rl:linear1,features1}.
A relevant  observation made early on is that the features can often be defined using a simple DL grammar from the domain predicates
\cite{martin:kr2000,fern:general}.
For example, if  $q(z)$ and $r(x,y)$ are two domain predicates of arities 1 and 2 respectively, one can define new unary predicates $p_1(x)$ and $p_2(x)$
as $\exists y [q(y){\land}r(x,y)]$ and $\forall y \, r(x,y)$, and use the new unary predicates  to define new ones, etc. Unary predicates $p$  can be used to define  numerical features  $n_p$,  whose value
is the number of objects that satisfy $p$ in a state $s$, and  Boolean features $b_p$,
whose value is true when $n_p$ is greater than $0$ \cite{bonet:aaai2019}.

Interestingly, most variants of DLs are parts of a fragment of first-order logic (FOL)
known as $\FO_2$, that involves just two variables,  such as $x$ and $y$ above \cite{description-logics}.
In other words, DLs can express some FO-formulas that make use of two variables but not three.
The extension of $\FO_k$, with $k$ variables, with  counting quantifiers
$\exists^{\ge i}$ to express  that there are  at least $i$ \emph{different} objects that comply with a formula
is the logic $\C_k$.

The  relation between the features required for expressing general policies and value functions,
and  the finite-variable logics required to express such  features is  relevant
as it has been  shown recently  that \emph{guarded} $\C_2$ ($\GC_2$), which corresponds to a
standard description logic, provides a tight characterization of the expressive power of
(message passing) graph neural nets \cite{barcelo:gnn,grohe:gnn}. This  suggests  that GNNs can be used
to learn general policies using the domain predicates without having to generate a pool
of $\C_2$ features by assuming some fixed grammar and a bound on the complexity of the features.
\emph{This is the hypothesis that we  explore in this work by focusing on the problem of learning general value functions that are optimal.}

\section{Value Functions for Tractable Tasks}

We  consider \emph{optimal value functions} for a number of tasks and domains,
selected mostly from  \citeay{nir:ecai2012}, where they are shown to be
solvable, optimally, in polynomial time, suggesting that a compact optimal
value function may exist.
All features are defined in terms of the domain predicates,
and for simplicity, they are all  Boolean. Predicates $p_G$ refer to predicates
$p$ evaluated in the goal (goal predicates); i.e., while an  atom like  $\pon(a,b)$ is true or false in
a state, the atom $\pgoal{\pon}(a,b)$ is true in a state iff it is true in the goal of the instance \cite{martin:kr2000}.

In these domains, the optimal value functions are linear and expressed as sums
$V^*(s)=\sum_{i=1}^{N} c_i\bracket{\varphi_i(s)}$, where $c_i$ is a constant and  $\bracket{\varphi_i}$ is the Iverson bracket that
evaluates to $1$ if feature $\varphi_i$ holds in $s$, and  else to  $0$. The  formulas $\varphi_i$ all belong to the
logic $\C_k$ for $k=2$, and in one case, for $k=3$.

In many domains, we need features that reflect the existence of paths of length $k$ from object $x$ to some object $y$ such that
some condition $T(y)$ holds, where objects are connected by ``edges'' $E(x,y)$.
This can be expressed in $\C_2$ as:
\begin{alignat*}{1}
  \ppath_0(x)   &= \ptarget(x) \,, \\
  \ppath_{k}(x) &= \exists y(\pedge(x,y) \land \ppath_{k-1}(y)) \,.
\end{alignat*}
The distance of a shortest path of length $k$ is then  captured by $\pspath_{k}(x)\,{=}\,\ppath_{k}(x)\,{\land}\,\neg\ppath_{k-1}(x)$,
while the existence of such path of length up to $N$ is captured by $\conn_N(x)\,{=}\,\ppath_0(x) \lor\,{\cdots}\, \lor \ppath_N(x)$.
The constant $N$ is related to a hyperparameter $L$ in the architecture, to be described below.
Notation $\pspath_k[\ptarget',\pedge']$ (resp.\ $\conn_N[\ptarget',\pedge']$) denotes $\pspath_{k}$ (resp.\ $\conn_N$) where $T/E$ are replaced by $T'/E'$ resp.
Additionally, $\ppos{\textsc{P}}{1}(x)\,{=}\,\exists y\,\textsc{P}(x, y)$ and $\ppos{\textsc{P}}{2}(x)\,{=}\,\exists y\,\textsc{P}(y, x)$
denote that $x$ appears as first and second argument of some atom.
Lastly, $\textsc{P}^{-1}(x, y)$ holds iff $\textsc{P}(y, x)$ holds.

\paragraph{Blocksworld: Clear and On.}

\inlinecite{GuptaNau92} showed that finding an optimal solution for Blocksworld is \np-hard.
We thus consider a tractable version where the goal $\pclear(x)$  is to clear a specific block $x$. The optimal value function decomposes as
\begin{alignat*}{1}
  \vstar    &= \bracket{\alpha \land H} + \textstyle\sum_{k=1}^N (2k-1)\bracket{B_k} \\
  \intertext{for features}
  \alpha    &= \exists x(\pgoal{\pclear}(x) \land \neg\pclear(x)) \,, \\
  H         &= \exists x\,\pholding(x) \,, \\
  B_{k}     &= \exists x(\pgoal{\pclear}(x) \land \eta_k(x)) \,, \\
  \eta_k(z) &= \pspath_k[\pclear, \pon^{-1}](z)
\end{alignat*}
where $\alpha$ holds in a state, if it is not a goal state,
$H$,  if holding some block, and $B_k$ (resp.\ $\eta_k(z)$), if there are $k$ blocks above $x$ (resp.\ $z$), $k \leq N$.

The other  version of Blocksworld corresponds  to instances with single-atom goals of the form $\pon(x,y)$ for some $x$ and $y$.
In this case, the optimal value function for problems with up to $N$ blocks above $x$ or $y$ decomposes as
\begin{alignat*}{1}
  \vstar &= \bracket{\alpha \land H} + 2\bracket{\alpha \land L} + \textstyle 2\sum_{k=1}^{N} k(\bracket{X_k} + \bracket{Y_k})
  \intertext{for features}
  \alpha &= \exists xy(\pgoal{\pon}(x,y) \land \neg\pon(x,y)), \\
  L      &= \exists xy(\pgoal{\pon}(x,y) \land (\neg\pclear(y) \lor \neg\pholding(x))), \\
  X_k    &= \exists xy(\pgoal{\pon}(x,y) \land \eta_k(x) \land \neg\conn_{N}[\ppos{\pgoal{\pon}}{2}, \pon](x)), \\
  Y_k    &= \exists xy(\pgoal{\pon}(x,y) \land \eta_k(y) \land \neg\conn_{N}[\ppos{\pgoal{\pon}}{1}, \pon](y))
\end{alignat*}
where
$L$ holds if $x$ is not held or cannot be stacked on $y$, and
$X_k$ (resp.\ $Y_k$) holds if there are $k$ blocks above $x$ (resp.\ $y$)
and $x$ (resp.\ $y$) is not above $y$ (resp.\ $x$).

\paragraph{Gripper.}

There is a  robot with two grippers, and a set of rooms containing balls.
While the goal is to move every ball to the correct room,
we consider the subproblem of moving a single ball, whose goal
is just $\pat(x,y)$ for some ball $x$ and room $y$.
The optimal value function is
\begin{alignat*}{1}
  \vstar &= \bracket{\alpha{\land}P} + 3\bracket{\alpha{\land}B} + \bracket{\alpha{\land}D} + 2\bracket{\alpha{\land}G} + \bracket{\alpha{\land}F}
\intertext{for features}
\alpha &= \exists xy (\pgoal{\pat}(x, y) \land \neg\pat(x, y)) \,, \\
  P &= \exists xy(\pat(x, y) \land \ppos{\pgoal{\pat}}{1}(x) \land \patrobby(y)) \,, \\
  B &= \exists xy(\pat(x, y) \land \ppos{\pgoal{\pat}}{1}(x) \land \neg \patrobby(y)) \,, \\
  D &= \exists xy(\pgoal{\pat}(x, y) \land \patrobby(y)) \,, \\
  G &= \exists xy(\pgoal{\pat}(x, y) \land \neg \patrobby(y)) \,, \\
  F &= \exists xy(\ppos{\pgoal{\pat}}{1}(x) \land \pat(x, y)\ \land \\
    &\qquad\qquad \neg\exists x(\pgripper(x) \land \pfree(x)))\,.
\end{alignat*}
where $\alpha$ holds when the goal is not achieved,
$P$ (resp.\ $B$) holds when Robby is (resp.\ is not) in the same room as the ball $x$ and Robby should pick up $x$ (resp.\ move to pick it up),
$D$ (resp.\ $G$) holds when Robby is (resp.\ is not) in the room $y$, and
$F$ holds when no gripper is free and Robby is not carrying the ball $x$.
It is important to note that when Robby picks up a ball, the ball is no longer in any room.

\paragraph{Transport.}

The task is to deliver packages using trucks of bounded capacity.
We consider a version where the goal is an atom $\pat{}(x, y)$ for package $x$ and destination $y$.
The optimal value function $V^*$ decomposes as the sum for $1\,{\leq}\,k\,{\leq}\,N$ of
\begin{alignat*}{1}
&(k+1)\bigl( \bracket{\alpha{\land}T_{k}} + \bracket{\alpha{\land}D_k} + \bracket{\alpha{\land}D'_\ell} \bigr) + \bracket{\alpha{\land}T_k{\land}F_k}
\end{alignat*}
for features
\begin{alignat*}{1} \alpha      &= \exists xy (\pgoal{\pat}(x, y) \land \neg\pat(x, y)) \,, \\
  \beta(x,y)  &= \pat(x, y) \land \ppos{\pgoal{\pat}}{1}(x) \,, \\
  \gamma(x,y) &= \pin(x, y) \land \ppos{\pgoal{\pat}}{1}(x) \,, \\
  L(y)        &= \exists x(\pvehicle(x) \land \pat(x, y)) \,, \\
  C(y)        &= \exists x(\pvehicle(x) \land \pat(x, y)\ \land \\
                &\quad\quad \exists y (\pcapacity(x, y)\ \land \\
                &\quad\quad\quad \exists x\,\pcapacitypredecessor(x, y))) \,, \\
  T_k         &= \exists xy (\beta(x, y) \land \pspath_{k}[L, \proad^{-1}](y)) \,, \\
  F_k         &= \exists xy (\beta(x, y) \land \neg \pspath_{k}[C, \proad^{-1}](y)) \,, \\
  D_k         &= \exists xy (\beta(x, y) \land \pspath_{k}[\ppos{\pgoal{\pat}}{2}, \proad](y)) \,, \\
  D'_k        &= \exists xy (\gamma(x, y)\ \land \\
                &\quad\quad \exists x (\pat(y, x) \land \pspath_{k}[\ppos{\pgoal{\pat}}{2}, \proad](x)))
\end{alignat*}
where $\alpha$ holds iff the goal is not achieved, $T_k$ determines the distance
to the closest truck, $T_k \land F_k$ determines if all closest trucks are full and need to drop a package,
and $D_k$ (resp.\ $D'_\ell$) determines the distance from the package (resp.\ truck the package is in) to the destination.

\paragraph{Rovers.}

Multiple rovers equipped with different capabilities (soil analysis, etc)
must perform experiments and send results back to lander.
A simple version where just the soil at some specific location must be sampled is considered.

A feature for identifying the closest available and capable rover to sample soil is needed.
Since each rover has its own map, shortest-path on different graphs must
be considered:
\begin{alignat*}{1}
  \ppath_0(r,x)  &= \patsoilsample(x) \,, \\
  \ppath_k(r,x)  &= \exists y (\pcantraverse(r, x, y) \land \ppath_{k-1}(r, y)) \,, \\
  \pspath_k(r,x) &= \pat(r, x) \land \ppath_{k}(r, x) \land \neg \ppath_{k-1}(r, x) \,.
\end{alignat*}
In addition to find the distance to the closest capable rover to sample the soil $R_k$,
features are need  to decide if such rover is full $F_k$,  if the soil has not been sampled $S$,  and if the goal has not been achieved $\alpha$.
Also needed are Boolean features $L_k$ to express the distance from the soil or rover
with the sample to some location where data can be sent to lander.
The optimal value function decomposes as:
\begin{alignat*}{1}
  \vstar &= \bracket{\alpha} + \bracket{S} + \textstyle\sum_{k=1}^N (k \bracket{R_k} + \bracket{F_k} + k\bracket{L_k}) \,.
\end{alignat*}
The formulas $\pspath_k(r,x)$ enter into the definition of the features $R_k$.
Since these formulas involve 3 variables, the features used
to decompose $\vstar$ do not belong to $\C_2$ or $\GC_2$.

\paragraph{Visitall.}

The task is to find a path that starts at an  initial vertex and visits  all vertices in a given  graph.
The simplified version involves a single target vertex to be visited.
For graphs with up to $N$ vertices, the optimal value function is
\begin{alignat*}{1}
  \vstar &= \textstyle \sum_{k=1}^{N} k \bracket{\alpha \land D_k}
  \intertext{for features}
  \alpha &= \exists x(\pgoal{\pvisited}(x) \land \neg\pvisited(x)) \,, \\
  D_k    &= \exists x(\patrobot(x) \land \pspath_{k}[\pgoal{\pvisited}, \pconnected](x)).
\end{alignat*}

\paragraph{Other Domains.}

Logistics, Miconic, Parking-behind and Parking-curb, and Satellite are also considered in the experiments.
We do not have space to discuss them in detail; however, the goals of all these problems are single atoms and optimal plan lengths are
bounded by (small) constants (also in Gripper). On the other hand, for the two versions of Blocksworld, Transport,
Rovers and Visitall the length of optimal plans is not bounded.

\section{Learning The Value Functions}

We turn to the problem of learning these value functions
using GNNs directly from the domain predicates.
For this, we review  GNNs, their logic, and
the GNN architecture used.

\subsection{Graph Neural Networks}

GNNs represent trainable, parametric functions over graphs \cite{gori:gnn,book:gnn}.
We focus on aggregate-combine GNNs (AC-GNNs)  \cite{barcelo:gnn,grohe:gnn} with $L$
layers that are specified with aggregate functions $\agg_i$, combination functions $\comb_i$, and a  classification function $\CLS$.
On input graph $G$, a  GNN maintains a state (vector) $\boldsymbol{x}_v\in\reals^k$
for each vertex $v\in V(G)$, and computation consists of updating these  states
throughout $L$ stages,  with  $\boldsymbol{x}^{(i)}_v$ denoting the states after stage $i$.
The parameter $k$ is the dimension of the node state or  embedding. The computation model for AC-GNNs corresponds to updates \begin{alignat*}{1}
  \boldsymbol{x}^{(i)}_v\ :=\
    \comb_i\bigl( \boldsymbol{x}^{(i-1)}_v, \agg_i\bigl( \multiset{ \boldsymbol{x}^{(i-1)}_w \,|\, w\in N_G(v) } \bigr) \bigr)
\end{alignat*}
where $N_G(v)$ is the set of neighbors for vertex $v$ in $G$, and
$\multiset{\ldots}$ denotes a multiset (i.e., unordered set whose
elements are associated with multiplicities).
That is, at stage $i$, each vertex $v$ receives the state of its neighbors
which are then aggregated, and the result combined with the current
state $\boldsymbol{x}^{(i-1)}_v$ to produce the next state $\boldsymbol{x}^{(i)}_v$.
The fact that $\agg_i$ maps multisets of states into real vectors
means that it does not depend on the source of the received messages.
GNNs are used  for node or graph classification. In the first case,
after the final stage, the  node $v$ is classified into the class
$\CLS(\boldsymbol{x}^{(L)}_v)$ determined by a  function $\CLS:\reals^k\rightarrow\{0,1\}$.
In the second case, the $\CLS$ function maps the multiset $\multiset{\boldsymbol{x}^{(L)}_v\,|\,v\in V(G)}$
into a single, scalar output; an operation  referred to as a \emph{readout}.

In our case, GNNs map planning states $s$ into real values $V(s)$. However,
the atoms in a planning state do not induce a graph (or hypergraph), but the
more subtle relational structure.
Therefore, we adapt below the GNN architecture to deal with relational structures.
In any case, the functions involved in the mapping from inputs to outputs can be linear
or non-linear, and they are all trainable; in the supervised case,
by minimizing a loss function defined over a training set given by pairs
$\tup{s,V^*(s)}$, where all the states $s$ (sets of atoms) come from
instances of different size but over a common planning domain and
common set of goal predicates.

\subsection{The Logic of GNNs}

The expressive power of AC-GNNs has been recently studied in relation to
decidable fragments of first-order logic \cite{barcelo:gnn,grohe:gnn}.
For this, it is convenient to consider the more general vertex-colored graphs $G$ and to assume that the AC-GNNs  for such graphs
initialize the embeddings of the vertices $\boldsymbol{x}^{(0)}_v$
to a  one-hot encoding of the vertex  colors.  One of the first crisp
results for  node classification is that  if the Weisfeiler-Lehman (WL) procedure,
a well-known coloring algorithm   that provides  a sound but incomplete
test for graph isomorphism \cite{wl:1968}, assigns the same color
to two nodes in a graph, then every AC-GNN classifier will map the
two nodes into the same class \cite{xu:gnn,morris:gnn}.

This result has been extended in two ways: one, where the WL procedure is
replaced by the logic $\C_2$, making use of a seminal result relating the
two \cite{immerman:c2-wl}, and the second, where  the characterization
of the expressive power of AC-GNNs is made tight, describing not just
what they can compute, but also what they cannot \cite{barcelo:gnn}.
For this, the logical formulas considered are those that involve equality and
two types of predicates: a binary  edge $E(x,y)$ predicate representing
the edges in the graph, and unary predicates $c_i(x)$ representing
the color of vertices. A (Boolean)  node  classifier can be expressed then as
a logical formula $\varphi(x)$ over these predicates with a  single free  variable $x$.
The question is what is the relation between
the node classifiers that can be captured in an AC-GNNs and those that can
be described logically.

The logical classifiers that can be captured by AC-GNNs are fully characterized in terms of
\emph{graded modal logic}  $\GC_2$, which is equivalent in expressive power to the standard
description logic  $\cal ALCQ$ \cite{barcelo:gnn}.
$\GC_2$ is the class of all formulas in $\C_2$ in which each quantified variable is \emph{guarded} by the edge relation;
e.g., $\psi(x)=\exists y\,[E(x,y) \land \mathsf{blue}(y)]$ that holds when $x$
has a blue neighbor.
The main result is:

\begin{theorem}[\citeauthor{barcelo:gnn}, \citeyear{barcelo:gnn}]
  \label{thm:1}
  A logical classifier is captured by AC-GNNs {if and only if}
  it can be expressed in  graded modal logic  ($\GC_2$), or equivalently,
  in the description logic $\cal ALCQ$.
\end{theorem}

Moreover,  each $\GC_2$ classifier can be  captured by a simple and homogeneous AC-GNN;
i.e., with linear combinators, and combinators and aggregators that are  identical across all layers.
There is a similar result for $\C_2$ classifiers, but this  requires an  slightly  modified  version of  AC-GNNs, called ACR-GNNs,
where  the combination function for each vertex $v$ is extended to take  an  extra
argument given by an aggregation of the  states for {all  vertices} in the graph:

\begin{theorem}[\citeauthor{barcelo:gnn}, \citeyear{barcelo:gnn}]
  \label{thm:2}
  Logical node classifiers in $\C_2$ are captured by simple and homogeneous ACR-GNNs. \end{theorem}

\subsection{GNNs for Relational Structures}

In as much as truth valuations give meaning to propositional formulas,
relational structures give meaning to first-order formulas. In the
case of planning states, the induced relational structures only have
relations, and do not involve constants nor functions.
Thus, a relational structure  $\R=(\A,R_1^\A,\ldots,R_m^\A)$ consists of
a domain of interpretation $\A$ and relations $R_i^\A$ of arity $k_i$ that stand for sets of $k_i$-tuples from $\A$.
In the relational structure defined by a graph, $\A$ is given by the vertices
and there is a single relation $R^\A$ given by the edges.
In the structure defined by a planning state $s$, $\A$ is given by the set of objects
in the instance, and $R_i^\A$  is the set of object tuples that satisfy the predicate $R_i$ in $s$.

Our modification of GNNs to handle relational structures is inspired by the one
introduced by \citeay{grohe:max-csp} for solving Max-CSP problems where all relations
are assumed to be binary and thus any such Max-CSP instance maps straightforwardly to a
directed graph. A major difference though is that our architecture does not make
use of LSTMs \cite{schmidhuber:lstm}.

For dealing with relations of any arity, the computation maintains states $\boldsymbol{s}^{(i)}_o$ for each object $o\in\A$
and proceeds in stages $i=1,\ldots,L$, where each atom $p=R(o_1,\ldots,o_n)$
computes messages $\boldsymbol{m}_{p,o_i}$ that are sent  to each  object $o_i$.  Each object $o$ then aggregates the incoming messages $\boldsymbol{m}_{p,o}$ from
the atoms $p$ that mention $o$, and combines this  aggregation with the current state $\boldsymbol{s}^{(i-1)}_o$
to produce the next state $\boldsymbol{s}^{(i)}_o$.
The final state (object)  vectors are passed through a neural net, aggregated,
and the result passed again to a final network to produce a
single output vector $\boldsymbol{v}$ of dimension $q$.
For relational structures that capture a state $s$,
the output $\boldsymbol{v}$  is aimed to approximate the scalar function $V^*(s)$,
and hence the output dimension  is $q=1$.

The architecture shown in Algorithm~\ref{alg:architecture} uses one
feed-forward neural net  $\mlp_R$ for each relational symbol $R$ (domain and goal predicate),
one such net $\mlp_U$ as a combination function, and
two nets $\mlp_1$ and $\mlp_2$ for constructing the final output $\boldsymbol{v}$.\footnote{Another major difference is that the messages $\boldsymbol{m}_{p,o}$ sent
  to objects, line 4 in Alg.~\ref{alg:architecture}, are computed with MLPs whereas in the
  architecture of \citeay{grohe:max-csp} the messages are computed with linear transforms.
}
All MLPs consists of a dense layer with a ReLU activation function,
followed by a dense layer with a linear activation function.
For the aggregation function $\agg$, we use either
sum or smooth maximum (implemented as LogSumExp). The trainable parameters are thus the trainable parameters in the MLPs,
while the hyperparameters  are the embedding dimension $k$, the
output dimension $q$, and the number of stages $L$.
The initial embeddings  $\boldsymbol{s}^{(0)}_o$  are  obtained by
concatenating the zero vector $\mathbf{0}$ and a random vector $\mathcal{N}(0,1)$, each of dimension $k/2$ \cite{Abboud21,Sato21}.

The parameters of the network  are learned by stochastic gradient
descent by minimizing the loss $\L(\R,\boldsymbol{\ell})=\|\boldsymbol{v}-\boldsymbol{\ell}\|_1$
from training data $\{(\R_i,\boldsymbol{\ell}_i)\}_i$. In our setting, the relational structures
$\R_i$ encode (the atoms that are true in)  the states  $s$, and the target value $\boldsymbol{\ell}_i$
for $s$ is $V^*(s)$.

\begin{algorithm}[t]\footnotesize
  \SetAlgoLined
\KwIn{Relational struct. $\R=(\A,R_1,\ldots,R_m)$ \ [states $s$)]}
  \KwOut{$\boldsymbol{v} \in \reals^q$ of dimension $q$ \ [value $V(s)$]}
  \BlankLine
  \tcp{\footnotesize Partial random initialization}
  $\boldsymbol{s}^{(0)}_{o} \sim \mathbf{0}^{k/2}\mathcal{N}(0, 1)^{k/2}$ for each object $o \in \A$\; \label{gnn:init}
  \For{$i \in \{1, \dots, L\}$}
  {
    \For{atom $p := R(o_1, \dots, o_n)$ with $\bar o\in R$}
    {
      \tcp{\footnotesize Generate messages $p \rightarrow o_j$}
              $(\boldsymbol{m}_{p,o_j})_j := \mlp_R(\boldsymbol{s}^{(i-1)}_{o_1}, \ldots, \boldsymbol{s}^{(i-1)}_{o_n})$\;\label{gnn:message}
    }
    \For{$o \in O$}
    {
      \tcp{\footnotesize Aggregate messages and update}
              $\boldsymbol{s}^{(i)}_{o} := \mlp_U\bigl( \boldsymbol{s}^{(i-1)}_{o}, \agg(\multiset{ \boldsymbol{m}_{p,o} \,|\, o\in p })\bigr)$\; \label{gnn:update}
    }
  }
  \tcp{\footnotesize Final Readout} \label{gnn:readout}
  $\boldsymbol{v} := \mlp_2\bigl(\sum_{o \in \A}\mlp_1(\boldsymbol{s}^{L}_o)\bigr)$
  \caption{General architecture (trainable, parametric function) that maps  relational structures $\R =(\A,R^\A_1,\ldots,R^\A_m)$
    into   vector $\boldsymbol{v}$. In our setting, $\R$ encodes the states $s$,  and $\boldsymbol{v}$ approximates $V^*(s)$.
    Atoms $p(o_1, \ldots, o_n)$ true in the input send messages to the objects $o_i$ in $p$,
    and objects $o$ aggregate all messages received and update their state  $\boldsymbol{s}^{(i)}_o$.
  }
  \label{alg:architecture}
\end{algorithm}

\section{Experiments}

We now evaluate if models (neural nets) can be trained and used as policies
in the domains and tasks considered above.
We first describe how states are sampled and labeled, then the experimental setup, and finally, the results.\footnote{Code and data: \url{https://doi.org/10.5281/zenodo.6353140}}

\subsubsection{Data.}

For a set of instances, we sample and label states for each as follows. First, we perform a single random walk $s_1, \dots, s_n$ from the initial state.
Then, for each $1{\leq}\,i\,{\leq}\,n$, we construct a planning problem with initial state $s_i$,
and find an optimal plan $s'_1, \dots, s'_m$ with A$^{*}$ using the admissible $h_{\textit{max}}$ heuristic \cite{bonet:aij-hsp}.
For each $1\,{\leq}\,j\,{\leq}\,m$, we add the pair $\tup{s'_j, m - j}$ to the dataset,
up to $40,000$ such pairs, balancing the number of states per label (distance).
The value of $n$ is set to produce that many pairs if possible.

\subsubsection{Setup.}

\begin{table}[t]
  \centering
  \begin{tabular}{@{\extracolsep{4pt}}lccc@{}}
    \toprule
Domain         &     Train & Validation &       Test \\
    \midrule
    Blocks-clear   &    [2, 9] &   [10, 11] &   [12, 17] \\
    Blocks-on      &    [2, 9] &   [10, 11] &   [12, 17] \\
    Gripper        &  [10, 18] &   [20, 22] &   [24, 48] \\
    Logistics      &  [17, 24] &   [31, 31] &   [31, 39] \\
    Miconic        &   [5, 26] &   [29, 35] &   [38, 92] \\
    Parking-behind &  [21, 27] &   [30, 30] &   [30, 36] \\
    Parking-curb   &  [21, 27] &   [30, 30] &   [30, 36] \\
    Rovers         &  [15, 52] &   [53, 62] &  [67, 116] \\
    Satellite      &  [14, 41] &   [47, 59] &  [50, 103] \\
    Transport      &  [14, 39] &   [38, 43] &   [41, 77] \\
    Visitall       & [27, 102] & [102, 146] & [171, 326] \\
    \bottomrule
  \end{tabular}
  \caption{Number of objects in the problems in the training, validation and test datasets;
    e.g., each problem for Miconic in the validation set has a number of objects in $[29,35]$.
  }
  \label{tbl:experiments:sizes}
\end{table}

The hyperparameters $k$ and $L$ are set to $32$ and $30$,
respectively; $k$ affects the number of features per object, but also training speed and memory usage.
The domain with the most predicates is Rovers with $32$ predicates, so the value for $k$ ensure
that at least one feature (scalar) per predicate is possible.
Our architecture can find shortest paths of length up to $2L$.
In the experiments, we evaluate nets with sum- and (smooth) max-aggregation denoted by \modelsum and \modelmax, respectively.
The architecture is implemented in PyTorch \cite{PyTorch} and each net is trained with
NVIDIA A100 GPUs for up to $12$ hours.
\modelsum is trained with L1 regularization set to $0.0001$, and no regularization for \modelmax (resulted in the lowest loss on the validation set).
Training is done with Adam \cite{Kingma15} with a learning rate of $0.0002$.

Table~\ref{tbl:experiments:sizes} shows the number of objects for the problems in the training, validation and test datasets.
We trained $5$ networks for each domain, and for each training session, the net with the best validation loss at the end of each epoch is selected.
Among the $5$ trained nets, the final net is the one with the best validation loss.
For the learned $V$ function, we run the policy $\pi_V$, selecting from each non-goal state $s$, the successor $s'$ with least $V$-value,
breaking ties by selecting the first such successor.
This is repeated for at most $100$ steps, or until a goal state is reached.
In the latter case, the problem is solved, and if the number of steps is minimal
(verified with A* and $h_{\textit{max}}$), \emph{the problem is counted as solved optimally.}

\begin{table}[t]
  \centering
  \begin{tabular}{@{\extracolsep{-3pt}}lccccc@{}}
    \toprule
    & & \multicolumn{2}{c}{\modelsum{}} & \multicolumn{2}{c}{\modelmax{}} \\
      \cmidrule(lr){3-4} \cmidrule(l){5-6}
    Domain (\#)         & L    &   Opt.     &     Sub. & Opt.       &    Sub. \\ \midrule
    Blocks-clear (11)   & 82   &   {\bf 11} &  0       & {\bf 11}   & 0       \\
    Blocks-on (11)      & 150  &   {\bf 11} &  0       & {\bf 11}   & 0       \\
    Gripper (39)        & 117  &   31       &  8       & {\bf 39}   & 0       \\
    Logistics (8)       & 48   &   5        &  3       & {\bf 8}    & 0       \\
    Miconic (95)        & 378  &   {\bf 95} &  0       & {\bf 95}   & 0       \\
    Parking-behind (32) & 77   &   {\bf 32} &  0       & {\bf 32}   & 0       \\
    Parking-curb (32)   & 101  &   7        &  12      & {\bf 32}   & 0       \\
    Rovers (26)         & 111  &   0        &  4       & {\bf 20}   & 6       \\
    Satellite (20)      & 97   &   {\bf 20} &  0       & {\bf 20}   & 0       \\
    Transport (20)      & 208  &   18       &  1       & {\bf 20}   & 0       \\
    Visitall (12)       & 93   &   {\bf 12} &  0       & {\bf 12}   & 0       \\ \midrule
    % Total (306)         & 1,462 & 242 (79\%) & 28 (9\%) & 300 (98\%) & 6 (2\%) \\
    Total (306)         & 1,462 & 242  & 28  & 300  & 6   \\
                        &       & (79\%) & (9\%) & (98\%) & (2\%) \\
    \bottomrule
  \end{tabular}
  \caption{Number of problems in test set solved optimally, suboptimally, or not solved at all with policy $\pi_V$ for learned $V$,
    when aggregation is done by \textsc{sum} or \textsc{max}. Total number of problems (\#) shown in parenthesis. Tasks and domains from Section 4.
    L is the sum of all optimal plan lengths.
  }
  \label{tbl:experiments:coverage}
\end{table}

\subsubsection{Results.}

As it is shown in Table~\ref{tbl:experiments:coverage}, the value functions
learned with \modelmax yield policies that \emph{solve all of the 306  test instances,
98\% of them optimally}. The 6 instances not solved optimally are all in Rovers, that as shown above,
requires $\C_3$ features.
This is a pretty impressive result that shows that deep nets can produce very crisp results.
In our case, it means that the \modelmax nets \emph{deliver policies that do not make a single mistake in
the plans of  300 test problems, and this means, practically no errors in the  1,462 intermediate decisions
made in the construction of these plans}. Notice that this is different than  simply measuring ``coverage''
(number of problems solved) where policies are allowed to make mistakes, if they are not fatal, and
typically ``noise'' is introduced to prevent being trapped in cycles. In terms of the aggregation
functions, the performance of \modelsum is not as good as \modelmax.
The  theory does not help us to understand this  difference, but it has been noted that
max-aggregation is better suited for discrete decisions and tasks that involve shortest paths
\cite{VelickovicEtal20}.

\section{Understanding the Learned Features}

We also tested if the learned features in the trained models can be understood
in terms of the hand-crafted features used in our analysis of the domains.
For this, let $y$ be the  vector of $n$ features based on our formal analysis, where distance features SP
are treated as numerical features. The readout function consists of a sequence of layers: (1) ReLU; (2) linear; (3) summation; (4) ReLU; and (5) linear.
Let $x$ be the concatenation of all intermediate feature vectors after aggregation in the readout function, i.e., the results of layer (3), (4) and (5).
Finally, let $y'=xA+b$ be a linear function of $x$ optimized such that  the linear coefficients  $A$ and $b$
minimize the loss  $\mathcal{L'}(y', y) = \sum_{i=1}^{n}|y'_i-y_i|$.
If this is loss is zero or very small, it means that the learned features encode a linear transformation of the hand-crafted features.

\begin{table}
\centering
    \begin{tabular}{@{}lccc@{}}
    \toprule
    Domain             & \#  & Train $\mathcal{L'}$ & Test $\mathcal{L'}$ \\ \midrule
    Blocks-clear       & $2$ & $0.12$               & $0.16$              \\
    Blocks-on          & $6$ & $0.88$               & $0.96$              \\
    Blocks-on-$\Sigma$ & $5$ & $0.17$               & $0.23$              \\
    Gripper            & $5$ & $0.04$               & $0.11$              \\
    Transport          & $4$ & $0.71$               & $1.13$              \\
    Transport-$\Sigma$ & $3$ & $0.30$               & $0.48$              \\
    Visitall           & $1$ & $0.06$               & $1.91$              \\ \bottomrule
    \end{tabular}
    \caption{Total loss $\mathcal{L'}$ of hand-crafted features over the train and test set, and the number of such features.
        Features are taken from the analysis of each domain (Section~4).
        Domains with $\Sigma$ replaces two numerical features by their sum.
}
    \label{tbl:learnedfeatures:loss}
\end{table}

Table~\ref{tbl:learnedfeatures:loss} shows the loss on the test set, after $A$ and $b$ are optimized on the training set.
In Visitall, this loss is largest among the domains considered (those for which $V^*$ was given in compact form), $1.91$,
and yet the optimal coverage is $100\%$, meaning that the distances are ordered well but not linearly.
The loss for Blocks-on and Transport over the training set is roughly $0.8$ and this suggests that
the networks do not learn one or more of the hand-crafted features well, although it turns out that
they learn a suitable aggregation of them.
The features for Blocks-on-$\Sigma$ in the Table~\ref{tbl:learnedfeatures:loss}, replace the two numerical features induced by $X$ and $Y$ in Blocks-on by their sum, and the same is done for Transport-$\Sigma$ for $D$ and $D'$.
The training and test losses then drop to roughly $20\%$ of the previous loss
in Blocks-on, and to $40\%$ in Transport, implying that these features are learned instead. \section{Understanding the Limitations}

The neural network  does not approximate well the optimal value function in Rovers, which is the only domain
where the optimal policy does not generalize $100\%$ with max aggregation. The problem is
that optimal policies for Rovers  require  $\C_3$ features that cannot be computed with standard GNNs.
Interestingly, the analysis reveals that this limitation is not due to the presence of multiple rovers, but to multiple rovers with their \emph{own maps}.
For illustrating this, we designed a simplified Rovers domain called Vacuum: an assortment of robot vacuums that have to clean a  specific spot.
The predicates of this domain are \pat/2, \pdirty/1, and \padjacent/3, and each robot $r$ can clean
a  location $x$ and move to an adjacent location $y$ if $\padjacent(r, x, y)$.
We consider three different versions: {Vacuum-R} with at most $5$ robots, {Vacuum-M} where all robots share the same traversal map, and {Vacuum} with no restrictions.
We generated $20$ problems of each version of Vacuum and ensured that optimal plan lengths vary from approximately 3-8 for the training set,
6-9 for the validation set, and 6-12 for the test set.
The number of problems solved optimally by \modelmax is $1$ for Vaccum, $4$ for Vaccum-R, and $20$ for Vaccum-M.
The only version with $100\%$ generalization (or close) is Vacuum-M, which is
precisely the version of the domain where there are $\C_2$ features for deciding the length of shortest paths.
The $r$ argument in $\padjacent(r, x, y)$ is indeed redundant, and if $\padjacent'$ denotes the
resulting binary predicate, the optimal value function for Vacuum-M decomposes as
\begin{alignat*}{1}
  \vstar = \textstyle\sum_{k=1}^N (k + 1) \bracket{D_k}
\end{alignat*}
for $D_k = \exists x (\pgoal{\pdirty}(x) \land \pspath_k[\ppos{\pat}{2}, \padjacent'^{-1}](x))$, which is a
$\C_2$ feature.

\section{Related Work}

\subsubsection{Neuro Symbolic AI.} Many proposals have been advanced for integrating symbolic and DL approaches
due to  limitations  and opacity  of pure data-based approaches
\cite{josh,luc:neuro,neuro-symb}.
Our integration combines domain predicates, that can potentially be learned \cite{asai:fol,bonet:ecai2020,ivan:kr2021},
builds on the correspondences between finite variable logics and GNNs \cite{barcelo:gnn,grohe:gnn}, and
modifies the architecture for Max-CSPs.
Interestingly, recent GNN methods can compute more general functions
that are not limited to those defined on the  $\C_2$ features associated with DLs logics only
\cite{Abboud21}.
\subsubsection{General Policies.} The problem of learning general  policies has been addressed  using combinatorial
approaches where the symbolic domains are given \cite{khardon:action,martin:kr2000,bonet:aaai2019,frances:aaai2021},
DL approaches where the domains are given too   \cite{sylvie:asnet,mausam:dl},
and DRL approaches that do not make use  of prior knowledge about  the structure of either  domains or states
\cite{sid:sokoban,babyAI,minigrid:amigo}. This work  is a  step  to bring the first two approaches together along with their potential benefits.

\subsubsection{General Value Policies.} It is known since the 1950s  that a value function  $V$ defines a policy $\pi_V$ which is optimal if $V$ is optimal  \cite{bellman:dp,bertsekas:dp,sutton:book}. Linear value functions have been
particularly important in RL until the advent of deep RL methods that dispense with the need for hand-crafted features
\cite{dqn,intro-drl}. In classical planning, linear value functions have been used  under the name of ``potential heuristics'' \cite{potential-heuristics},
where   the features are   conjunctions of atoms, and   ``generalized potential heuristics'' \cite{frances:ijcai2019},
where   the ($\C_2$)  features are the Boolean and numerical features based on DLs \cite{bonet:aaai2019}.
A ``descending and dead-end avoiding potential function'' $V$ represents indeed a value function $V$ that defines a greedy policy $\pi_V$
that solves a problem. The proposed  learning method provides  crisp experimental evidence that
generalized value functions with $\C_2$ features can be computed without having to \emph{explicate the pool of features and without having to
assume a linear combination.} Our focus on optimal value functions is methodological:
it allows for supervised learning with $V^*$ targets, and  a crisp evaluation (no single mistake allowed in the execution of plans).
The same learning approach  can be used in stochastic MDPs where the targets $V^*$  represent optimal expected costs to the goal.
Also, due to the correspondence between $\C_2$ features and GNNs,
the same architecture can be used for learning value functions without supervision \cite{frances:ijcai2019},
possibly using RL methods.

\section{Summary}

Previous works  have shown that general policies and value functions  for many classical  planning  domains can be expressed
in terms of a pool of features that is obtained from the domain predicates using a DL grammar, and learned without supervision
using  combinatorial  solvers. In this work, we have exploited  the relations between DLs and the decidable  fragment $\C_2$  of
FOL, and between GNNs and $\C_2$, to approach  a similar   problem  (optimal policies and value functions)
but avoiding the grammar, the complexity bounds, and  the combinatorial  solvers that have been replaced
by more robust and scalable  deep learning engines.

Other authors  have addressed the problem of learning general policies using GNNs and GNN-like architectures given the domain descriptions. What distinguishes our approach
is that our deep learning architecture is  simple and general; a modification of a GNN architecture  introduced for solving a completely different task:
Max-CSPs over binary constraints \cite{grohe:max-csp}. We also have  a logical characterization of what are we trying to learn  and
we have  used  it to understand  the scope of the computational model (power and limits), and what is actually learned. Recent extensions of GNN learning, however,
suggest  that (value)  functions of features that are  more complex than those associated with DLs could be learned effectively as well.

\section*{Acknowledgments}

This research was partially supported by the European Research Council (ERC), Grant No.\ 885107, and by project TAILOR, Grant No.\ 952215, both funded by the EU Horizon research and innovation programme.
This work was partially supported by the Wallenberg AI, Autonomous Systems and Software Program (WASP) funded by the Knut and Alice Wallenberg Foundation.
The computations were enabled by the supercomputing resource Berzelius provided by National Supercomputer Centre at Link\"{o}ping University and the Knut and Alice Wallenberg foundation.

\bibliography{paper}

\end{document}